\documentclass[letterpaper, 10 pt, conference]{ieeeconf}  % Comment this line out if you need a4paper

\IEEEoverridecommandlockouts                              % This command is only needed if 
\usepackage{epsfig} % for postscript graphics files
\usepackage{times} % assumes new font selection scheme installed
\usepackage{amsmath} % assumes amsmath package installed
\usepackage{amssymb}  % assumes amsmath package installed
\usepackage{graphicx}
\usepackage{hyperref}
\usepackage{xcolor}
\usepackage{multirow}
\usepackage{siunitx}
\usepackage{array}
\usepackage{diagbox}
\usepackage{adjustbox}
\usepackage{booktabs}
\usepackage{todonotes}
\usepackage{pifont}
\usepackage{amssymb}
\usepackage{url}
\usepackage{colortbl}
\usepackage{overpic}

\newcolumntype{P}[1]{>{\centering\arraybackslash}p{#1}}

\renewcommand{\baselinestretch}{0.985}

\newcolumntype{R}[2]{%
    >{\adjustbox{angle=#1,lap=\width-(#2)}\bgroup}%
    l%
    <{\egroup}%
}

\usepackage[font=small]{caption}
\title{\LARGE \bf
Improving Deep Dynamics Models for Autonomous Vehicles with \\ Multimodal Latent Mapping  of Surfaces}

\author{Johan Vertens$^{*}$ \and Nicolai Dorka$^{*}$ \and Tim Welschehold \and Michael Thompson \and Wolfram Burgard% <-this % stops a space
\thanks{$^{*}$These authors contributed equally. Johan Vertens, Nicolai Dorka, and Tim Welschehold are with the University of Freiburg, Germany. Michael Thompson is with Toyota Research Institute, Los Altos, USA. Wolfram
Burgard is with the University of Technology, Nuremberg, Germany. Corresponding author: {\tt\small vertensj@informatik.uni-freiburg.de}}
}

\begin{document}

\maketitle
\thispagestyle{empty}
\pagestyle{empty}

%%%%%%%%%%%%%%%%%%%%%%%%%%%%%%%%%%%%%%%%%%%%%%%%%%%%%%%%%%%%%%%%%%%%%%%%%%%%%%%%
\begin{abstract}
The safe deployment of autonomous vehicles relies on their ability to effectively react to environmental changes.
This can require maneuvering on varying surfaces which is still a difficult problem, especially for slippery terrains.
To address this issue we propose a new approach that learns a surface-aware dynamics model by conditioning it on a latent variable vector storing surface information about the current location.
A latent mapper is trained to update these latent variables during inference from multiple modalities on every traversal of the corresponding locations and stores them in a map.
By training everything end-to-end with the loss of the dynamics model, we enforce the latent mapper to learn an update rule for the latent map that is useful for the subsequent dynamics model.
We implement and evaluate our approach on a real miniature electric car.
The results show that the latent map is updated to allow more accurate predictions of the dynamics model compared to a model without this information.
 We further show that by using this model, the driving performance can be improved on varying and challenging surfaces. 
\end{abstract}

\section{INTRODUCTION}
\label{sec:introduction}

In recent years autonomous cars have become reliable enough to be deployed in the real world~\cite{teslaautopilot,waymoone}.
Nonetheless, they can currently operate safely only under limited conditions such as a mapped environment, good weather, or specific types of roads.
In the unstructuredness of the real world with unforeseeable situations, potentially caused by mistakes of other road users, the autonomous car might be required to drive at the edge of its ability under all environmental conditions in order to avoid accidents.
One such scenario is driving on varying surfaces which becomes especially challenging if their corresponding friction values differ a lot.
Maneuvers that are safe to execute on one surface can be dangerous or even impossible on a different surface.
For autonomous vehicles, this poses a safety-critical problem as the driving has to be adjusted according to the current surface and a wrong estimate about possible future trajectories can result in disastrous outcomes.
How to autonomously learn to predict features of a surface and the corresponding impact on the future trajectory without supervision by ground truth friction values for all surfaces is still an open research question.

\begin{figure}
\centering
\includegraphics[width=0.94\linewidth]{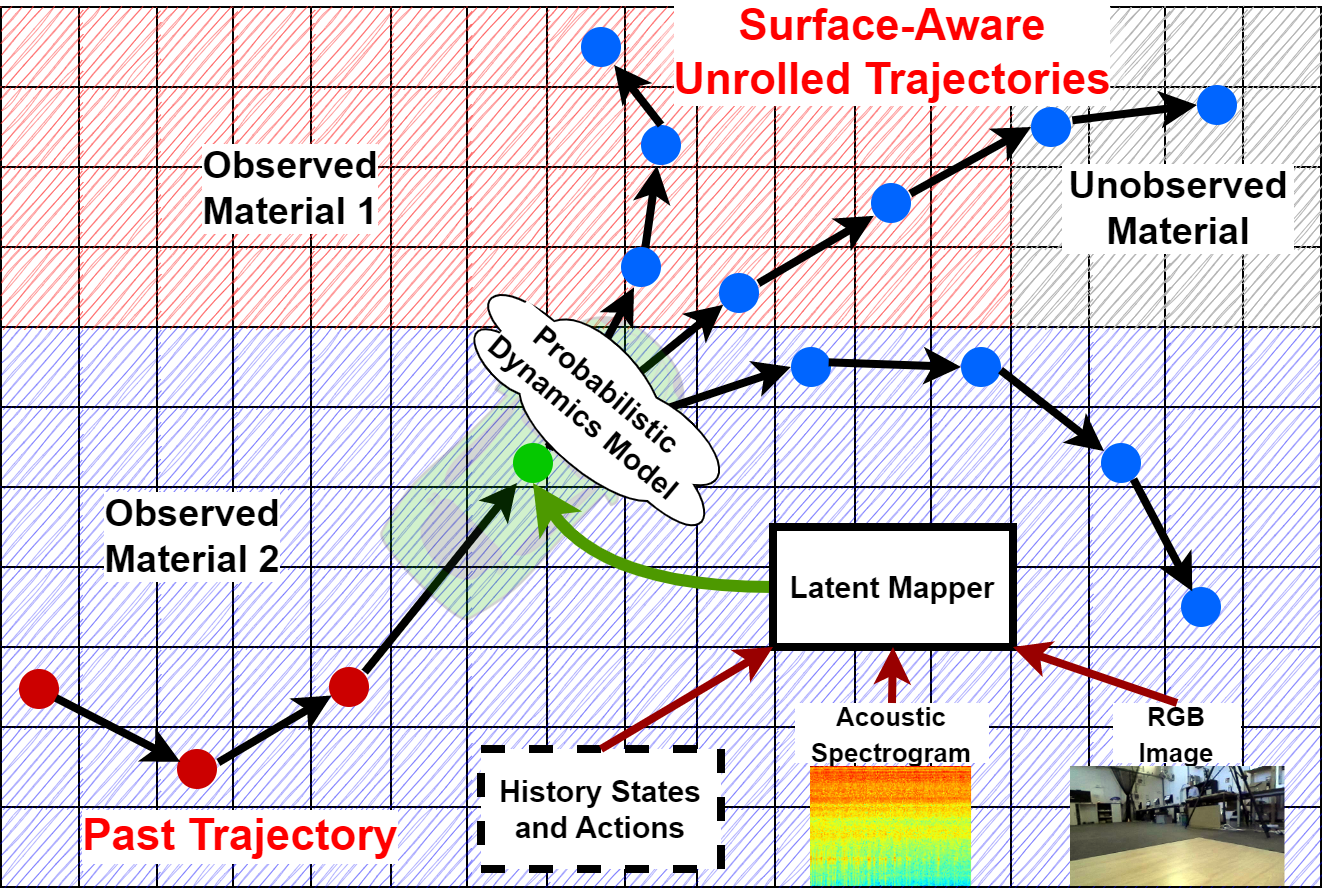}
\caption{
Overview of our approach.
During driving the latent mapper uses different modalities like sound and vision to update grid cell variables that represent information about the road and its properties at that location.  
The learned dynamics model receives the latent variable corresponding to its current location as additional input to allow for trajectory predictions that are aware of the road material.
}
\label{fig:cover}
\vspace{-0.55cm}
\end{figure}

Previous works allowing stable handling at the edge of controllability considered a single surface~\cite{cutler2016autonomous,1528936,8710588} or achieved steady state drifting but no cornering~\cite{Acosta2018TeachingAV}.
A typical approach in the case of varying surfaces is to use a separate terrain classifier~\cite{kalakrishnan2010fast,kolter2007hierarchical,thrun2006stanley}, which, however, assumes a fixed number of prespecified terrain classes.
This problem can be overcome by using conditional dynamics models~\cite{nagabandi2018learning} but the inferred information is not stored in a map and can hence not be used for later traversals of the same location.
Other works estimate the explicit coefficient of friction inside a physics model~\cite{8917024,huang2019calculation}. This, however, relies on the assumption that an accurate model can be identified, which is difficult for slippery surfaces.

We tackle this problem with a new approach that trains a dynamics model to include information from a learned and automatically updated latent map.
During inference, a latent mapper updates the map, such that the latent variable at a specific location stores valuable information about the surface material at that position. 
The dynamics model uses the latent variable for its current location as additional input to make accurate surface-dependent predictions of the next state.
Several traversals through a single location are given sequentially to the latent mapper to update the corresponding latent variable in an autoregressive fashion.
Since all model parts are differentiable, we can train the dynamics model and the latent mapper simultaneously from the same loss in a way that the latter learns to give useful surface information to the former.
Motivated by the fact that humans automatically use multiple cues to infer what driving style is appropriate for the surface they are driving on, the latent mapper receives multiple modalities like images and sound allowing it to learn a general representation of different surface patterns.

We implement and evaluate our method on a real miniature electric car.
The results show our approach is able to generate complex driving maneuvers on unknown and varying surfaces, showing the benefit of implicit terrain maps.
To summarize, our contributions are
\begin{itemize}
    \item a latent mapper trained to update a map with surface information from multiple modalities
    \item a surface-aware dynamics model using the latent map
    \item the implementation on a real electric car
    \item an extensive evaluation of the single components of our method and their combinations
    \item demonstrating the benefit of our latent mapping relative to a baseline without surface awareness in the real world.
\end{itemize}

\vspace{-0.3cm}
\section{RELATED WORK}
\label{sec:relatedworks}

\vspace{-0.1cm}
\subsection{Surface-Awareness}
\vspace{-0.1cm}
Terrain classification can be done in an automated way~\cite{zurn2020self} or from different modalities~\cite{valada2018deep}.
Several approaches use an explicit terrain classifier trained with human-specified labels and deploy a terrain-specific policy accordingly~\cite{kalakrishnan2010fast,kolter2007hierarchical,thrun2006stanley}.
More similar to our approach is the work of Nagabandi \textit{et~al.}~\cite{nagabandi2018learning} where additionally to state and action an image is given as input to the dynamics model such that it can infer the next state conditioned on the surface the robot is currently moving on.
Furthermore, meta-learning has been used for learning to adapt the dynamics model parameters during inference to a given environment including different terrains~\cite{nagabandi2018meta}.
Our approach is different in that it updates the surface embedding during inference allowing it in principle to learn the adaption to any new surface and to further store the information in a map.

\vspace{-0.1cm}
\subsection{Explicit Estimation of the Coefficient of Friction}
\vspace{-0.1cm}
One can compute possible trajectories using the coefficient of friction between the tires and the road. However, the coefficient depends on all kinds of conditions and has to be estimated from real data. 
For example, Huang \textit{et~al.}~\cite{huang2019calculation} estimate the coefficient with a limited-memory adaptive extended Kalman Filter to reduce the effect of outdated measurements on filtering. Multiple surveys~\cite{khaleghian2017technical, wang2022tire} cover these topics.
Deep neural networks can use ultrasound as input to classify the road surface from which the coefficient of friction is derived~\cite{kim2021road}.
For autonomous race cars driving several laps on a track with an inhomogeneous surface covering has been proposed to estimate the coefficient of friction and store it in a map~\cite{8917024}. 
All of these approaches have in common that they use a physical dynamics model for which they estimate the corresponding parameters. Our approach is more flexible in that it can learn any dynamics model without the need to model the underlying physics. Thus, our work is most comparable to other methods that learn dynamics models solely from data. Our focus is to improve ML-based methods by introducing surface-awareness.

\vspace{-0.1cm}
\subsection{Dynamics Models with Latent Variables}
\vspace{-0.1cm}
Some works train a network for system identification from the most recent history to output a latent variable which is then given as additional input to a model-free RL policy~\cite{yu17preparing} or the dynamics model in model-based RL~\cite{lee2020context,zhou2018environment}.
In the meta-RL setting Rakelly \textit{et~al.}~\cite{rakelly2019efficient} train a network to produce a probabilistic context variable given as additional input to an off-policy RL algorithm to adapt to a given task.

Our approach differs in that it builds a map over the state space and learns to learn different context embeddings at different locations in the state space during inference.

\vspace{-0.1cm}
\section{TECHNICAL APPROACH}
\label{drift:sec:approach}
In our work, we employ a dynamics model that predicts a potential future state given an input action. Subsequently, the dynamics model is used to control a vehicle along pre-defined trajectories following a model predictive scheme.
Our approach does not model explicit parameters of a physical model nor does it take factors such as tire pressure or temperature into account. 
One could combine these aspects with out method but this is beyond the scope of this work.

In contrast to previous approaches, we propose a method that considers potential changes of the road material along the track, which may influence the dynamic behavior of the vehicle. 
Therefore, we first create a grid map that is defined in global coordinates and equally divides the space into quadratic cells. 
Once the vehicle traverses a cell of the grid map, a mapping neural network infers a latent vector that describes the local road surface. 
Thereby, the mapping network leverages various modalities that were recorded during the traversal of the cell, such as RGB images, acoustic spectrograms, history states, and history actions. 
Our dynamics model takes, aside from low-level state and action information, the latent vector corresponding to the current vehicle location as an input, such that it can leverage the additional mapped cues from multiple modalities to optimize the predictions of future states, making it surface-aware.
As the training of the mapping model is guided by the loss of the dynamics model predictions, we do not require any labels that associate the input modalities with specific surface characteristics.
Note, that in contrast to previous approaches, our work models the whole surface-aware dynamics as a neural network, which avoids assumptions or inductive biases for the created surface map.
Next, we unroll multiple trajectories using the dynamics model and sampling from the action space. 
Our system then employs a reward function and the cross entropy~\cite{de2005tutorial} method to score and select the trajectories that follow a reference path as fast and close as possible. 
Following a typical model predictive scheme, we execute the first action of the resulting plan and restart the process.
In the following we describe each component of our system in more detail, followed by an explanation of the loss functions and training procedure.

\vspace{-0.1cm}
\subsection{Surface-Aware Probabilistic Dynamics Model Ensemble}
\vspace{-0.1cm}
\label{drift:sec:dyn_model}
Our dynamics model predicts the next output state $s^{\mathrm{out}}_{t+1}$, given the current input state $s^{\mathrm{in}}_{t}$ and the current action $a_t$. To this end, we distinguish between input states, which are the representation of the input to the dynamics model, and output states, which are the prediction of the dynamics model.
We define the input state as the concatenation of 3D linear velocities $\mathrm{vel}_l$, angular velocities $\mathrm{vel}_a$, linear accelerations $\mathrm{acc}_l$, angular accelerations $\mathrm{acc}_a$, and motor rpm : $s^{\mathrm{in}} = [\mathrm{vel}_l^T, \mathrm{vel}_a^T, \mathrm{acc}_l^T, \mathrm{acc}_a^T, \mathrm{rpm}]^T$. 
Additionally, we define the predicted output state of the dynamics model as estimated local changes in the x-position $\Delta p_x$, y-position $\Delta p_y$ and yaw angle $\Delta \gamma$, all velocities, and all accelerations as  $s^{\mathrm{out}} = [\Delta p_x, \Delta p_y, \Delta \gamma, \mathrm{vel}_l^T, \mathrm{vel}_a^T, \mathrm{acc}_l^T, \mathrm{acc}_a^T, \mathrm{rpm}]^T$.
The action is composed of the throttle $a_\mathrm{th}$ and steering command $a_\mathrm{st}$, such that $a=[a_\mathrm{th}, a_\mathrm{st}]$.
We make our model surface-aware by using additional latent vectors as input to the dynamics model. These latent vectors describe the local learned properties of the road.
Therefore, we propose to learn a latent map $L$ that is represented as a grid map where each quadratic cell $c$ holds a distribution over the latent vector. Here, we assume each entry of the map to be a $k_l$-dimensional multivariate normal distribution that is parametrized by $l_{\theta}^c$, corresponding to cell $c$. As the vector is learned implicitly, the number of dimensions $k_l$ represents a hyperparameter. 
To predict the next output state, we first sample a latent vector $l^c$ from the latent distribution $\mathcal{N}(l_{\theta_\mu}^c, {l_{\theta_{\sigma^2}}^c})$ at the cell corresponding to the current vehicle position. 
Following, we employ an ensemble of probabilistic dynamics models~\cite{chua2018deep} with parameters $\psi$ to predict the next state, while capturing model and data uncertainties. The input of this ensemble comprises the current input state, the current action, and the sampled latent vector, which we feed by simple concatenation. 
Thus, we model the Gaussian distribution of the next state as:
\vspace{-0.1cm}
\begin{equation}
    f_{\psi}(s_{t+1} \mid s_{t}, a_t) = \mathrm{Pr}(s_{t+1} \mid s_{t}, a_t, l^c;\psi).
    \vspace{-0.1cm}
\end{equation}
For clarity, we omitted the differentiation between the input and the output state.

\vspace{-0.1cm}
\subsection{Mapping Network}
\vspace{-0.1cm}
\label{drift:sec:mapping_network}
To estimate the latent map, we propose a novel neural network architecture that takes a variety of modalities as input.
In more detail, when the car traverses a cell $c$, we leverage an RGB image $I^c$, an acoustic spectrogram $S^c$, the history of states $H_s^c$, the history of actions $H_a^c$, and the previous estimate of the mean and variance of the latent vector $l_{\theta}^c$ that were recorded in the same cell $c$. The cues are then encoded into high-level features using respective encoders.
These features are then concatenated and passed to another MLP with two output heads, which predicts the mean and variance of the latent vector respectively.
As mentioned in Sec.~\ref{drift:sec:dyn_model}, the means and variances are then aggregated to a latent map $L$.
More formally, let $\phi$ be the parameters of the mapping model and $\Tilde{l_{\theta}^{c}}$ the updated parametrization of the Gaussian distribution of the latent vector in cell $c$. We define a latent update for the cell $c$ as: 
\vspace{-0.1cm}
\begin{equation}
\Tilde{l_{\theta}^{c}} = M_{\phi}(I^c, S^c, H_{s}^c, H_{a}^c, l_{\theta}^c),
\vspace{-0.1cm}
\end{equation}
Note, that since we input previous latent estimates $l_{\theta}^c$, the latent representation of the road surface is updated iteratively. 

While the RGB image may entail visual information of the road, the acoustic spectrograms capture direct tire-road interactions that are characteristic for specific materials. Additionally, from learning about the history of states and actions, our mapping model can infer ground patterns that lead to specific state sequences given the respective actions.

Particularly at inference time, estimating the road characteristics from only the low-level state and action history would most likely fail when the vehicle is at slow speeds or stands still.
To argue about the road surface, the dynamic behavior needs to differ across different road materials due to distinct friction characteristics. This, however, is only given in cases where the maximum frictional force~\cite{mu2003estimation} that is achieved is smaller than the force that is needed to sustain the vehicle track. Thus, to enable arguing about the road material, situations are required in which the car starts slipping.
While, for training purposes of the dynamics model, this data can be collected by an expert driver, uncontrolled slipping should be avoided during inference. 
To this end, our multimodal approach allows learning a mapping that associates visual or acoustic cues to a latent representation of the road without the requirement of slipping. As an example, our network can learn to associate a slippery surface with the visually shiny appearance of the road. Further, acoustic spectrograms can contain information about the road even under low velocities.

Consequently, we require examples of aggressive driving only during training, while during inference the road representation can be estimated under all conditions.
Furthermore, the employed modalities can complement each other if one modality lacks information due to visual occlusions, low lighting,  or when external acoustic events drown out important acoustic tire-road interactions.
We show in Sec.~\ref{drift:sec:results} that leveraging multimodal data yields high gains in state prediction performance.

\vspace{-0.1cm}
\subsection{Training of the models}
\vspace{-0.1cm}
\label{drift:sec:training}
We first collect a training dataset with dynamic examples of random driving in environments with spatially changing road materials. These driving examples include situations where the car slips.
To train our networks we propose a loss function that fulfills two requirements:
\begin{itemize}
    \item Unrolling of future states may lead to querying cells that have not been observed yet. In these cases, it should be possible to inform the dynamics model of zero knowledge of the surface material.
    \item The outputs of the mapping model should represent a spatial property of the road surface that is valid for any state prediction in the respective cell. 
\end{itemize}
To accomplish these requirements, we first group the individual recorded ground-truth state transitions according to the cell $c$ in which the transition was captured. Here, we denote the $n$-th state-transition in our dataset that occurred in the cell $c$ as $s^{c^n}_{t} \rightarrow  s^{c^n}_{t+1}$. Now, having a list of all state transitions that occurred in the same cell, we select $N$ random state transitions within a cell and define the loss for training the dynamics model as:
\vspace{-0.1cm}
\begin{equation}
\label{drift:eq:main_loss}
\mathcal{L}_d = \sum_{n=[0, 1, ... N]}(\mathcal{L}_g( f_{\psi}(s_{t+1}^{c^n} \mid s_{t}^{c^n}, a_t^{c^n}, l^{c^n} ),  \Bar{s^{c^n}_{t+1}})),
\vspace{-0.1cm}
\end{equation}
where $l^{c^0} = 0$ and for $n>0$:
\vspace{-0.1cm}
\begin{equation}
% l^{n+1} = M_{\phi}\left(I^{c^{n}}, S^{c^{n}}, H_{s}^{c^{n}}, H_{a}^{c^{n}}, l_{\theta}^{c^{n}}\right),
l^{c^{n+1}} = M_{\phi}\big(I^{c^{n}}, S^{c^{n}}, H_{s}^{c^{n}}, H_{a}^{c^{n}}, l_{\theta}^{c^{n}}\big),
\vspace{-0.05cm}
\end{equation}
and where $\mathcal{L}_g(\theta, \mathrm{target}) = \frac{1}{2} (\log \theta_{\sigma}^2 + \frac{(\theta_\mu - \mathrm{target})^2}{\theta_{\sigma}^2})$ is the Gaussian negative log-likelihood loss, and $\Bar{s^{c^n}_{t+1}}$ is the ground truth target state. The order of the traversals is irrelevant to our loss. Our loss represents the update scheme of the latent vectors, in which in the first iteration no knowledge of the surface is assumed. Thus, in the first iteration, a latent vector for the dynamics model is defined as a zero-vector, which forces the dynamics model to predict a future state distribution that is broad enough to cover all road materials properties that appear during training. This is particularly useful when unrolling state sequences over cells that have not been observed yet. In these cases, a conservative estimate of the state distribution is required as unobserved cells may entail any material or surface condition. 
In the following iterations of our loss function ($n >0$), the latent vector fed to the dynamics model is updated using our mapping model. In our loss function, the latent update is calculated based on the observed data of the previous traversal $n-1$, while the dynamics model predicts the state transition for the current traversal $n$ in the same cell $c$.  Thus, we ensure that the latent vectors are independent of the currently predicted state transition and represent a joint representation that improves prediction accuracy for all transitions in the respective cell.
Note that if the dynamics model would receive a latent vector generated by the mapping network using data recorded at the same time as the inputs of the dynamics model, the mapping model could directly contribute to the prediction of the next state rather than representing a spatial property of the track.
For our experiments, we set the number of selected state transitions to $N=3$.

\subsubsection{Three Stage Training}
We optimize our model using a three-stage approach. In the first stage, we optimize the dynamics model as well as the latent vectors but without training the mapping network. Instead, we optimize the latent vectors $l^c$ directly by backpropagating into them, treating the map as model parameters. We denote the directly optimized latent parameters as $\Bar{l^{c}} \in \Bar{L}$. In contrast to the limited information of the input of the mapping network at inference time, this has the advantage that the latent vectors can be thoroughly optimized over all batches of the dataset. 
We denote the resulting loss as:
\vspace{-0.1cm}
\begin{equation}
    \mathcal{L}_{d_{\mathrm{s1}}} = \sum_{n=[0, 1, ... N]}(\mathcal{L}_g( f_{\psi}(s_{t+1}^{c^n} \mid s_{t}^{c^n}, a_t^{c^n}, \Bar{l^{c}}),  \Bar{s^{c^n}_{t+1}}))
\vspace{-0.1cm}
\end{equation}
In contrast to the later training stages, we do not inject zero-vectors, while optimizing the latent vectors directly as we presented in Eq. \ref{drift:eq:main_loss}. Experiments have shown that the training becomes instable otherwise.

% To avoid agitated latent maps, we additionally add a smoothness term that minimizes the local gradient of the latent maps:
To avoid agitated latent maps, we add a smoothness term minimizing the local gradient of the latent maps:
\vspace{-0.15cm}
\begin{equation}
\mathcal{L}_s = \; \mid \mid \nabla \Bar{L} \mid \mid^2    
\vspace{-0.15cm}
\end{equation}
The overall loss being optimized in the first stage is simply a weighted sum of both loss functions:
\vspace{-0.15cm}
\begin{equation}
    \mathcal{L}_{\mathrm{s1}} = \mathcal{L}_{d_{\mathrm{s1}}} + \lambda \mathcal{L}_s,
\vspace{-0.15cm}
\end{equation}
where $\lambda$ denotes a weighting hyper-parameter that defines the strength of the smoothness term.

However, as the parameter map $\Bar{L}$ is optimized offline, it can not be employed in practical applications as the vehicle should be capable of driving through previously unseen environments. By leveraging our multimodal mapper, new environments should be observed on-the-fly avoiding this limitation.
Thus, in the second stage, we freeze the learned latent parameters $\Bar{L}$ and the dynamics model while optimizing the mapping network. In this stage, we guide the latent predictions from our mapping model, by optimizing the negative log-likelihood of the predicted latent vector distribution $l_{\theta}^{c}$ given the learned parameter corresponding to the same cell $\Bar{l^{c}}$.
Overall, the loss for the second stage of our training scheme is defined as:
\vspace{-0.15cm}
\begin{equation}
    \mathcal{L}_{\mathrm{s2}} = \sum_{n=[1, ... N]} \mathcal{L}_g(l_{\theta}^{c^n}, \Bar{l^{c^n}}).
\vspace{-0.15cm}
\end{equation}
In the last stage, we then freeze the mapping model and refine the dynamics model by optimizing Eq.~\ref{drift:eq:main_loss} and feed the estimate $l^c$ from the mapping network into the dynamics model instead of the previously used learned parameter $\Bar{l^{c}}$.

\begin{figure*}
\centering
\includegraphics[width=0.9\linewidth]{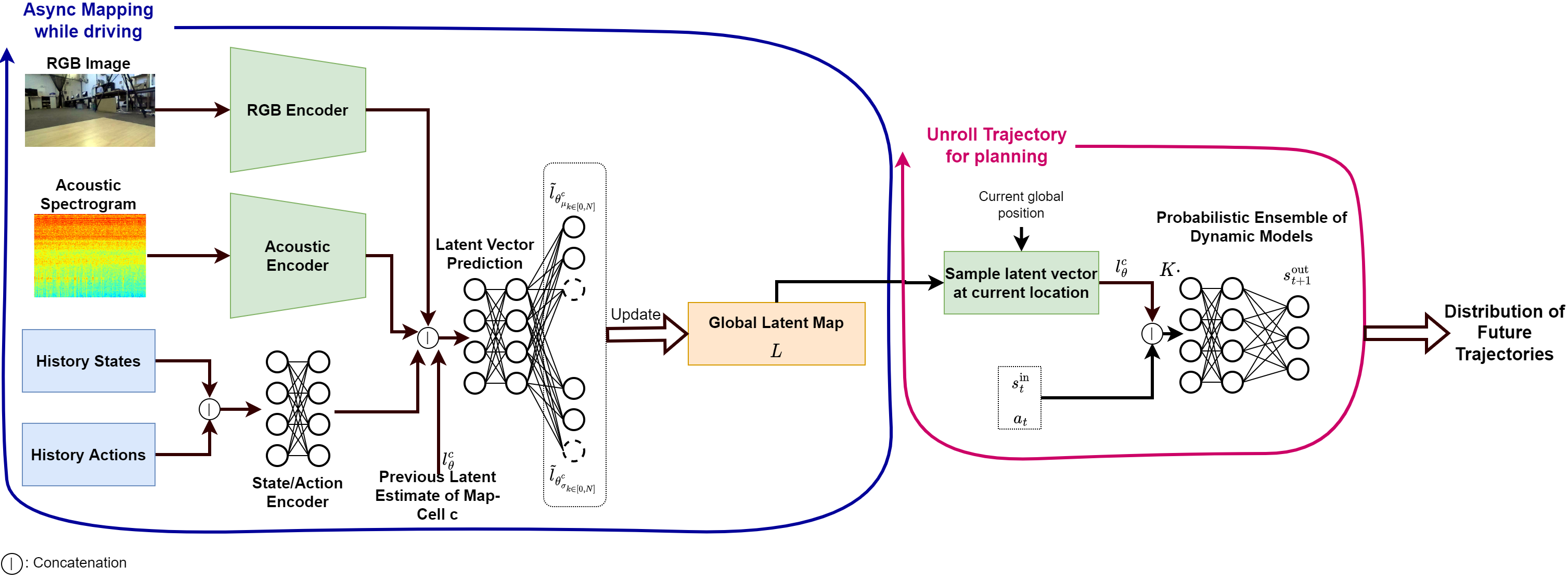}
\caption{In our approach the vehicle estimates a distribution of a latent representation of the road surface using different modalities, such as history state/action information, RGB images, and acoustic spectrograms. Thereby, all local latent distributions are collected in a global grid map $L$.  Once the vehicle traverses a cell $c$ of the grid map, the underlying latent distribution $l_{\theta}^c$ gets updated. For planning purposes, an ensemble of probabilistic dynamics models estimates the distribution of the future state $s_{t+1}$ given the actions $a_t$, the current input state $s_t$ and a sampled latent vector from the latent distribution of the grid-map-cell that corresponds to the input state $s_t$. As the dynamics model can be successively applied, a distribution of future trajectories can be obtained. As both mechanisms, latent mapping and state prediction can run independently from each other, the mapping is conducted asynchronously.}
\label{fig:architecture}
\vspace{-0.5cm}
\end{figure*}

\vspace{-0.1cm}
\subsection{Planning and Control}
\vspace{-0.1cm}
In order to follow a reference trajectory $T_r$, we follow a model predictive control approach. In detail, we use iCEM~\cite{pinneri2020sample}, which generates multiple future state sequences by sampling over the actions. 
The best sequences are selected using a pre-defined reward function and are refined for a specific amount of iterations. In contrast to the vanilla CEM~\cite{de2005tutorial}, iCEM provides significantly better sample efficiency and generates smoother trajectories due to enforced temporal consistency along the state sequences. These properties make the sampling-based planning real-time capable, which is a crucial requirement for high-speed autonomous driving.
\subsubsection{Trajectory Unrolling and Latent Sampling}
To generate candidate trajectories for the planning module, we start from the initial current state of the vehicle and unroll future state sequences by successively applying our dynamics model given a sequence of actions. We then accumulate all predicted local changes of the x-position $\Delta p_x$, y-position $\Delta p_y$, and yaw angle $\Delta \gamma$ to convert the state sequences into trajectories that are defined in the coordinate system of the initial state. 
Finally, we add the initial position of the vehicle to convert them into the global coordinate system. To sample multiple trajectories from our ensemble of dynamics models, we employ the \textit{TS1}-strategy \cite{chua2018deep}. In each iteration of iCEM we generate trajectories for $N_a$ distinct action sequences. As our ensemble of dynamics models predicts a single state transition at a time, we apply our dynamics model $h$ times for a trajectory with the length of $h$ time steps and the same number of actions. Further, we sample $k$ different state hypotheses for each individual action, effectively resulting in $N_a*k*h$ inferences passes of our dynamics model and $k*N_a$ trajectories. During trajectory generation, the latent vectors are sampled from the respective latent distribution for each individual state of the trajectories and leveraged for the next update using the dynamics model. Thus, intra-trajectory changes of the road materials are considered.
\subsubsection{Map Update}
We start with a map in which the parametrization of each cell is set to zero. As discussed in Sec.~\ref{drift:sec:training} this indicates zero knowledge about the map.
As the latent distribution in each cell can be updated asynchronously with respect to the prediction of the dynamics model, we run the mapping model and the controller including the dynamics model in separate threads.
This ensures that the dynamics model does not need to wait for the mapping inference to finish and allows for parallelization as we run the dynamics model on CPU and the mapping model on GPU.
\subsubsection{Reward Function}
To score all trajectory candidates, we propose a reward function that evaluates these in terms of different metrics.
The target trajectory is given by $x-y$ coordinate waypoints.
We measure the deviation of a given trajectory to the target with the cross-track error $R_{cte}$.
As we want the car to move as fast as possible we define a progress reward $R_p$ by adding up the length of all the line segments between waypoints the trajectory passes.
Further, to punish risky maneuvers and prevent shortcuts we define a binary boundary violation reward $R_b$ which is equal to $1$ if a specified boundary around the waypoints is exceeded and $0$ otherwise.
Lastly, to encourage smooth driving we define the reward $R_a$ as the absolute difference between the last executed throttle command and that of the first action of the trajectory.
The final reward for a trajectory is defined as
\vspace{-0.1cm}
\begin{equation}
    % R = w_p \cdot R_p - w_{cte} \cdot R_{cte} - w_a \cdot R_a + w_t \cdot R_t - w_b \cdot R_b,
    R = w_p \cdot R_p - w_{cte} \cdot R_{cte} - w_a \cdot R_a - w_b \cdot R_b,
\vspace{-0.1cm}
\end{equation}
where we set the weighting parameters to $w_p=40, w_{cte}=10,w_a=20, w_b=20000$.

\vspace{-0.1cm}
\section{DATASET}
\label{sec:dataset}
\begin{figure}
\setlength{\tabcolsep}{1pt}
\centering
\footnotesize
        \begin{tabular}{P{.2665\linewidth}P{.36\linewidth}P{.36\linewidth}}
        The Car & Without Map & With Map \\
        \includegraphics[width=\linewidth]{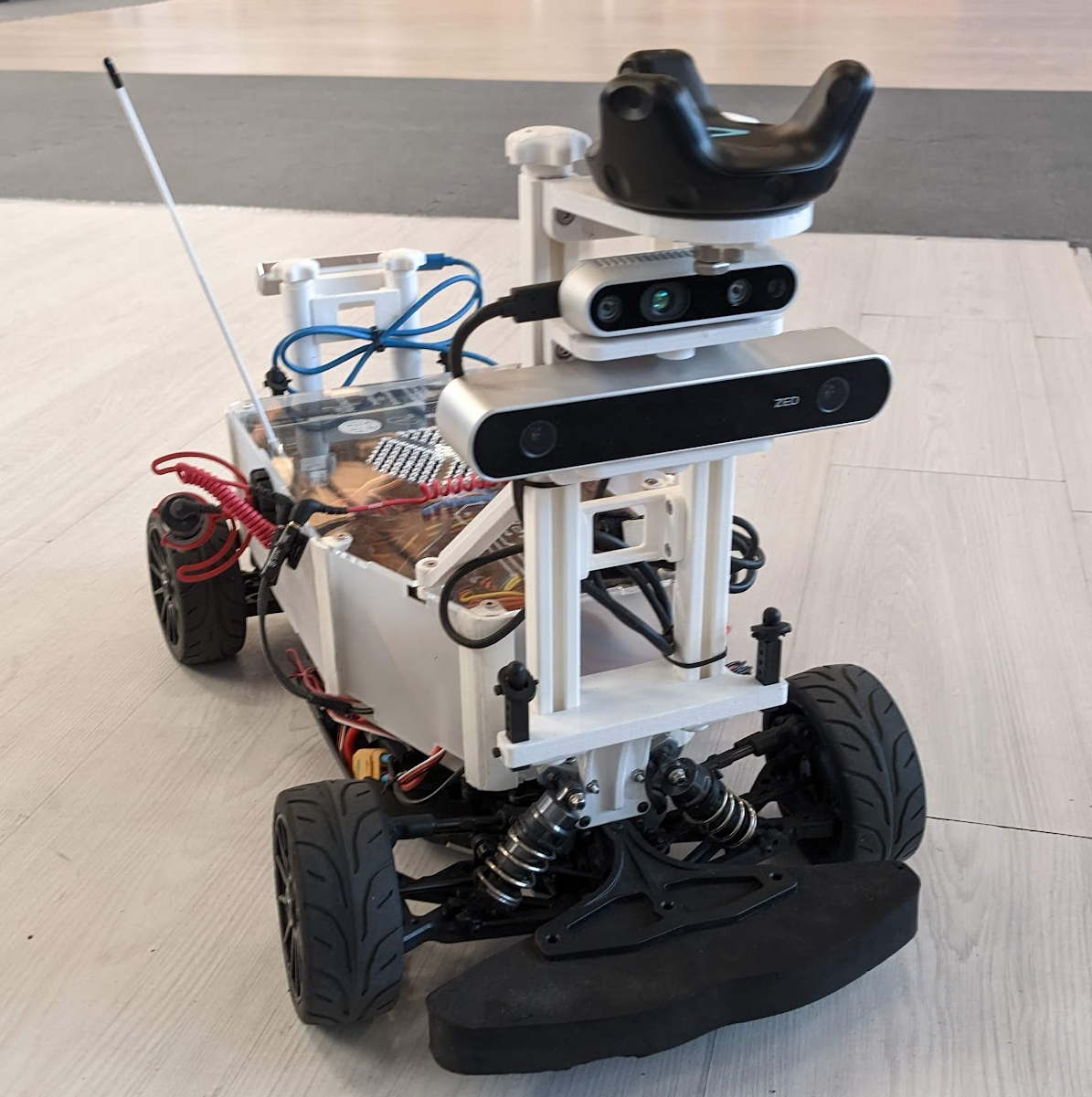} & \includegraphics[width=\linewidth]{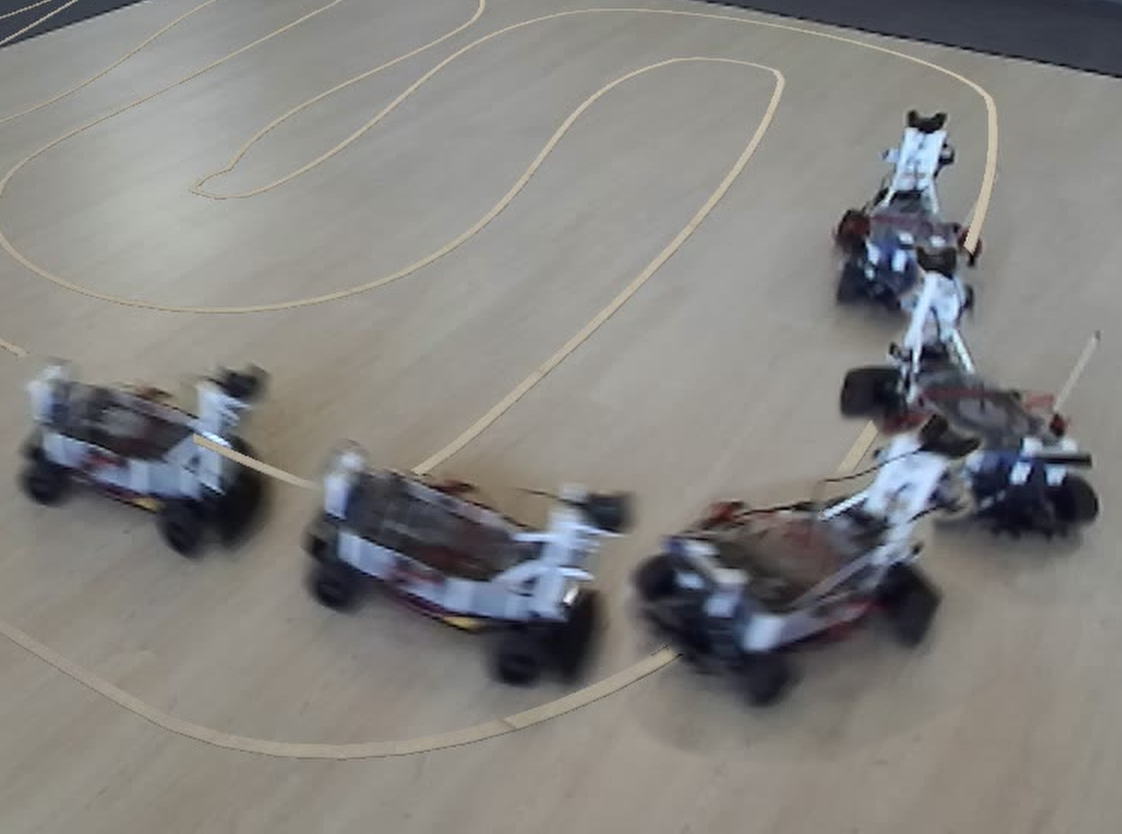} & \includegraphics[width=\linewidth]{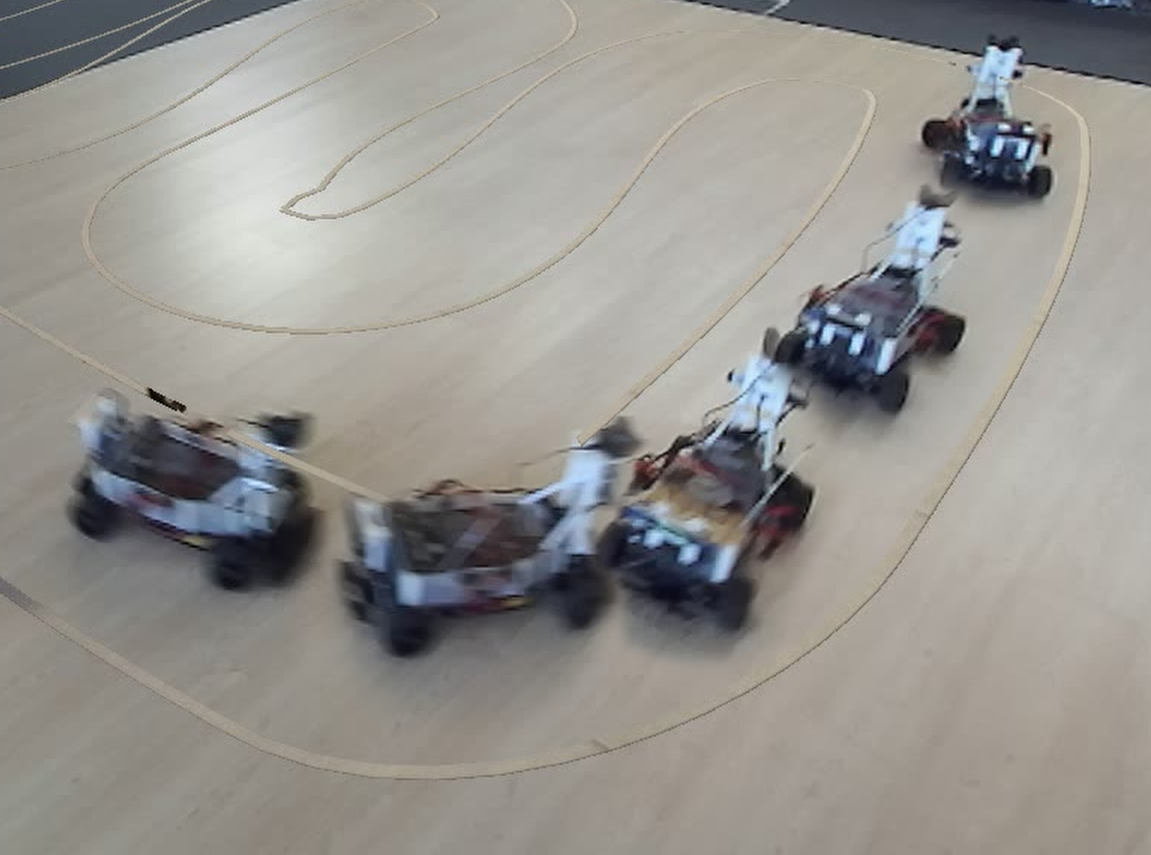} \\
    \end{tabular}
    \scriptsize  % smaller
    \caption{Left: Our experimental vehicle. The background shows the various surface materials over which the driving is conducted. Center and right: the driven path of the vehicle with and without map information.}
    \label{drift:fig:path_overlay} 
    \vspace{-0.5cm}
\end{figure}

To train and evaluate our approach we propose a novel dataset which we refer to as \textit{Dynamic FreiCar}. As an experimental vehicle, we employ a rear-driven miniature 1:8 scale car that is equipped with a computer, various sensors, and a high-torque electric motor.
To capture RGB images, we leverage a \textit{ZED} camera, while we use a \textit{Rode} compact microphone to record audio data. We further use \textit{Valve Lighthouses} to track the position of the car and gather ground-truth velocity and acceleration data.
Our dataset contains 70 minutes of expert driving along random trajectories that include drifting scenarios. All data is recorded at 100hz. 
We use two types of wood laminate and gym rubber mats for the different surface materials.
To avoid bumps at the transition we level out the different materials.
To ensure fair training and evaluation splits we create two maps for training and one map for evaluation by spatially rearranging the materials. Thus, as we train on multiple maps, we avoid overfitting to a specific map, which we validate in our experiments section. Figure \ref{drift:fig:path_overlay} shows our experimental vehicle and the driving environment.
Note that our approach does not model explicit friction values and we do not have the ground truth friction values for the different surfaces.

\vspace{-0.1cm}
\section{Implementation Details}
\vspace{-0.1cm}
We implement our approach on top of MBRL \cite{Pineda2021MBRL}, a pytorch-based~\cite{NEURIPS2019_9015} general framework for model-based reinforcement learning. We employ a ResNet-18 architecture for the image and spectrogram encoders respectively. For the base architecture of the dynamics model, we leverage the implementation of the Gaussian ensemble from MBRL.

We use a learning rate of $0.001$ for optimizing the parameters of the dynamics model and the map parameters, and a different learning rate of $0.0001$ for the mapping model. 
We train all models with a batch size of 96 across four \textit{RTX 2080 ti} GPUs for 500 epochs. For inference and running the model predictive controller, we use a Ryzen 5950X processor.

To extract the spectrograms from the acoustic data, we use an FFT with 257 bins and a hop length of 128. We extract the spectrogram over the last second of data and convert it to decibels afterward. The images and spectrograms are resized to $336$x$188$ before passing them to their respective encoders.

We further define the time span of a single state transition as $0.1$s and set the map resolution to $50$cm. As for the iCEM optimizer, we set $\beta = 4$ and $\gamma=1.3$. Furthermore, we use a planning horizon of $h=8$ steps, $32$ samples in each iteration, and set the number of iCEM iterations to 2.

\section{Experimental Results}
\label{drift:sec:results}
To assess the efficacy of our method we train on the training set of \textit{Dynamic FreiCar}.
\vspace{-0.05cm}
\subsection{Evaluation of the Dynamics Model}
\vspace{-0.05cm}
We evaluate the prediction accuracy of the dynamics model on the test set of our dataset. To generate a test set that represents practical use-cases, we leverage our method to drive 10 laps autonomously on track 3 (see Sec.~\ref{drift::planning_control}). We record all state transitions during this run and employ them as the test set for the following evaluation.
As unrolling long trajectories are practically important for planning and control, we introduce a metric that describes how well the predicted states align with the ground truth while considering uncertainty.
Thus, for all time steps in our dataset, we unroll a sequence of $N_s$ states using the future actions that have been carried out following the current state. To capture multiple hypotheses we repeat the unrolling process $100$ times for the same action sequence. Now, given $100$ hypotheses of the future trajectory, we compute the euclidean distance between each point of every trajectory hypothesis and the observed ground-truth trajectory that was driven. Finally, we compute the mean euclidean error.
More formally, let $D_s$ be the set of all observed states in the dataset, $H(s)$ all unrolled hypotheses starting in state $s$, and $\overline{T(s)_n}$ the n-th point of the observed ground-truth trajectory that starts in $s$. Then we define the metric as:
\vspace{-0.1cm}
\begin{equation}
    L2_{N_s} =\frac{1}{N_D N_H N_s}\sum_{s \in D_s} \sum_{h \in H(s)} \sum_{n=0}^{N_s} \mid \mid h_n - \overline{T(s)_n} \mid \mid_2 .
    \vspace{-0.1cm}
\end{equation}
where $N_D$ is the number of starting states in our dataset and $N_H=100$ is the number of rolled-out hypotheses.
We compute our evaluation metrics in the chronological order of the test dataset by iteratively updating the latent cues of the grid map as more data about the surface can be observed over time. This evaluation scheme strictly represents the practical deployment of the model, since the full map can not be leveraged at the beginning but builds up progressively.

We compare our approach against a baseline that does not employ any latent mapping (Ours-w/o map) similar to PETS~\cite{chua2018deep} and additionally present the performance of our approach when only a subset of the modalities are leveraged for the mapping model.
We denote a model that uses only images, spectrograms or history state/action information for latent mapping as \textit{Ours (I)}, \textit{Ours (A)} or \textit{Ours (S)} respectively. Furthermore, we show the results of a dynamics model that takes the ground-truth terrain types instead of the estimated latent vectors of the map as input. To encode the ground-truth surface we simply provide a scalar value to the model indicating on which terrain type it is operating. We denote this model as \textit{Ours (GT)}. We argue that this model should provide the highest accuracy as the terrain is known at all times.
As described in Sec.~\ref{sec:relatedworks}, the settings tackled in previous works deviate considerably from ours in terms of surface map representation and hence we refrain from comparing to them. In our work, we set the focus on evaluating the benefits of latent surface maps for dynamics models.

The results in Table \ref{drift:tab:quantitative} show that our multimodal latent mapper (\textit{Ours (AIS)}) significantly boosts prediction performance. 
By learning and mapping multimodal cues about the surface material, we reduce the error for $N_s=30$ from $0.462$ to $0.374$ corresponding to a reduction of $19\% $ over a model without the mapping model. In comparison to the model with the ground-truth terrain types as input (\textit{Ours (GT))}, our model is on par with respect to the $L2_{10}$ metric and only $1.9\%$ and $4.2\%$ worse for the $L2_{20}$ and $L2_{30}$ metrics respectively. As the performance of the model with ground-truth terrain types is expected to be an upper limit, one can argue that our model effectively predicts the surface information needed to improve prediction performance. 

Further, one can observe that the dynamics model performs best when using a mapping model that takes all modalities as input. 
Leveraging only the history of state/action information yields an $L2_{30}$ error of $0.422$, 
only acoustic yields $0.397$, and only visual cues yields $0.383$.
An interesting observation is that the combination of the history of states/actions and acoustic cues reaches a very low $L2_{30}$ error of $0.393$, even though images are not employed in this case. In contrast to RGB images, acoustic spectrograms do not contain information on the spatial location of the car. Thus, the image-free version of our mapping model is less prone to overfitting. Although we could not observe overfitting of the mapping model during our experiments, this could ease training in more large-scale scenarios.
Further, some of the improvement of using all modalities over just using the image might come from the fact that the front-facing camera only captures the surface farther in front of the current position, which might result in poor image-terrain associations at surface transitions.

While we used a ten-dimensional latent vector ($k_l=10$) for the previously explained models, we additionally evaluate models that leverage all modalities and employ $1$ and $5$ elements. These models are denoted as \textit{Ours (AIS,$k_l=1$)} and \textit{Ours (AIS,$k_l=5$)} respectively. Here, one can note that estimating ten-dimensional latent vectors additionally improves the $L2_{30}$ metric by $10\%$ over a model that employs a single dimension. We explain this effect by the thesis that overparametrization of the surface information eases the training procedure and helps avoiding local minima.  Estimating more than ten latent dimensions did not show further performance gains during our experiments.

\begin{table}
\footnotesize  % original
% \scriptsize  % smaller
\centering
\begin{tabular}{c|c|c|c}
Model & $L2_{10}$ $\downarrow$ & $L2_{20}$ $\downarrow$ & $L2_{30}$ $\downarrow$\\
\noalign{\smallskip}\hline\hline\noalign{\smallskip}
Ours-w/o map & $0.094$ & $0.262$ & $0.462$ \\
Ours (S) & $0.090$ & $0.242$ & $0.422$ \\
Ours (A) & $0.085$ & $0.228$ & $0.397$ \\
Ours (I) & $0.081$ & $0.218$ & $0.383$ \\
Ours (AS) & $0.082$ & $0.225$ & $0.393$ \\
Ours (AIS-$k_l$=1) & $0.087$ & $0.237$ & $0.418$ \\
Ours (AIS-$k_l$=5) & $0.080$ & $0.223$ & $0.400$ \\
Ours (AIS) & $\mathbf{0.079}$ & $\mathbf{0.212}$ & $\mathbf{0.374}$ \\
\noalign{\smallskip}\hline\noalign{\smallskip}
Ours (GT) & $\mathbf{0.079}$ & $\mathbf{0.208}$ & $\mathbf{0.359}$\\
\hline\noalign{\smallskip}
\end{tabular}
\caption{Quantitative evaluation of the prediction accuracy of our proposed dynamics network. We present comparisons to a model that is not using a latent map as input and to a model that leverages a map that is optimized offline as parameters.}
\label{drift:tab:quantitative}
\vspace{-0.3cm}
\end{table}

\vspace{-0.05cm}
\subsection{Qualitative Latent Maps}
\begin{figure}
\centering
\footnotesize
\setlength{\tabcolsep}{0.1cm}% for the horiz padding
    \begin{tabular}{P{2.5cm}P{2.5cm}P{2.5cm}}
        GT & Ours-Mean(AIS) & Ours-Var(AIS) \\
        \begin{overpic}[width=1.3\linewidth,tics=10, angle =90]{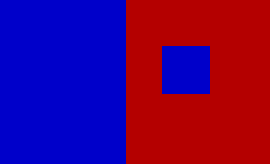}
 \put (5,2) {\textcolor{red}{\small \textbf{gt-layout}}}
\end{overpic}  & \begin{overpic}[width=1.3\linewidth,tics=10, angle =90]{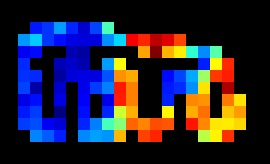}
 \put (5,2) {\textcolor{red}{\small \textbf{mean}}}
\end{overpic}  & \begin{overpic}[width=1.3\linewidth,tics=10, angle =90]{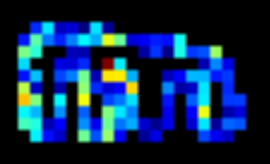}
 \put (5,2) {\textcolor{red}{\small \textbf{variance}}}
\end{overpic}  \\
    \end{tabular}
    \scriptsize  % smaller
    \caption{Qualitative results of aggregated predicted latent vectors. It can be observed that the predicted maps are representing different materials well.}
    \label{drift:fig:qualitative_maps} 
    \vspace{-0.6cm}
\end{figure}

\vspace{-0.05cm}
To investigate the learned latent maps in more detail, we visualize the learned map using a color-coding and $k_l=10$.
A principal component analysis (PCA) projects the multi-dimensional latent vector to a single scalar. We create the latent vectors using all available data of the test-split ($10$ laps) of our dataset. 
The updates of the latent vectors are conducted in chronological order of the split. 
Fig.~\ref{drift:fig:qualitative_maps} shows the learned map, the predicted variance of the latent estimate, and the corresponding ground-truth layout. %The offline optimized parameter map is not shown, as it is only optimized in the training set with different environments.
The experiment clearly reveals that our latent embedding correlates with the real-world layout of the test environment. Thus, we can validate that our mapping model predicts spatial cues of the track surface. Further, we observe that the predicted variance of the latent cues correlates with the locations of transitions between different materials. This follows our intuition as the observed data corresponding to a single cell of the grid map can contain information about multiple materials, since the cell can potentially stretch over material transitions. As the test split entails a distinct arrangement of the surface materials not seen during training, we can confirm that our mapping model generalizes over the map structure.

\subsection{Real-World Planning and Control}
\vspace{-0.03cm}
\label{drift::planning_control}
To evaluate our surface-aware dynamics model for practical applications, we employ our model predictive controller for autonomous racing.
The experiment is designed to investigate the effect of the dynamics model being surface-aware or not under otherwise same conditions. The dynamics model could be further improved by taking other features such as tire pressure and temperature into account. However, the car does not have sensors for this and these are orthogonal improvements that are beyond the scope of this work.
We conduct the experiments on three diverse single-lane maps varying significantly in curvature.
Track 1 and 3 have three different surfaces with highly varying friction values while Track 2 is completely on a slippery surface.
The accompanying video shows all maps as well as the resulting driving of our approach.
% The maps as well as the resulting driving of our approach can be seen in the accompanying video.
We quantify the driving performance in terms of three metrics computed over 30 laps.
In more detail, we present the average lap time, the average cross-track error (CTE), and the average lane-boundary violation score.
The lap time is a useful performance indicator as wrong dynamics predictions either lead to overly careful driving or to overly aggressive maneuvers on slippery surfaces which results in deviations from the racing line or emergency braking. The other two performance indicators focus on measuring wrong predictions in the context of overly aggressive driving.
The results in Tab.~\ref{drift:tab:quantitative_driving} suggest, that our surface-aware dynamics model significantly improves all stated metrics.
Our surface-aware model achieves significant lower lap times.
Furthermore, it yields reduced cross-track error and significantly reduces the violation of lane boundaries. Thus, our approach increases driving safety in challenging environments.

\begin{table}
\footnotesize  % original
% \scriptsize  % smaller
\centering
\begin{tabular}{c|c|c|c}
Model &  Lap Time $\downarrow$  & CTE $\downarrow$ & Bd. Violations $\downarrow$ \\
\noalign{\smallskip}\hline\hline\noalign{\smallskip}
Track 1 & \multicolumn{3}{c}{\raisebox{-.5\height}{\includegraphics[height=1.3cm]{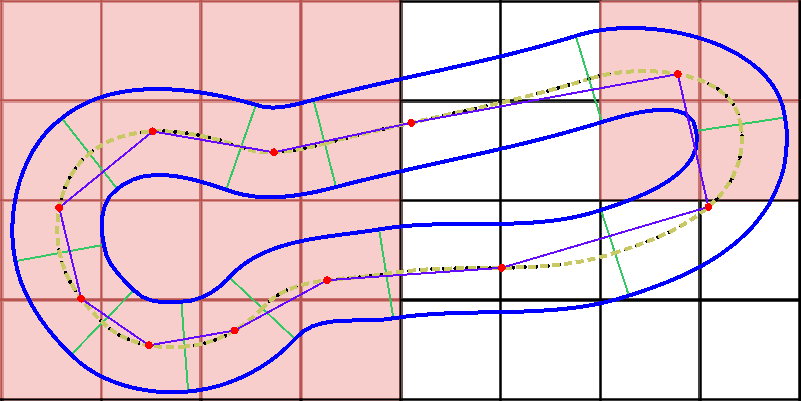}}}\\
\hline\hline\noalign{\smallskip}
Ours w/o map & $8.17 \pm 0.59$ & $27.63 \pm 7.95$ & $4.82 \pm 8.64$ \\
Ours (AIS)  & $\mathbf{7.35 \pm 0.42}$ & $\mathbf{26.34 \pm 6.12}$ & $\mathbf{2.79 \pm 5.98}$ \\
\hline\noalign{\smallskip}
Track 2 & \multicolumn{3}{c}{\raisebox{-.5\height}{\includegraphics[height=1.3cm]{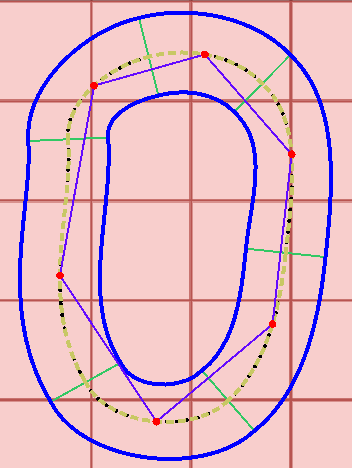}}}\\
\hline\hline\noalign{\smallskip}
Ours w/o map & $5.47 \pm 0.34$ & $\mathbf{14.57 \pm 4.92}$ & $1.39 \pm 4.89$ \\
Ours (AIS) & $\mathbf{4.89 \pm 0.26}$ & $17.33 \pm 3.48$ & $\mathbf{0.30 \pm 1.29}$ \\
\hline\noalign{\smallskip}
Track 3 & \multicolumn{3}{c}{\raisebox{-.5\height}{\includegraphics[height=1.3cm]{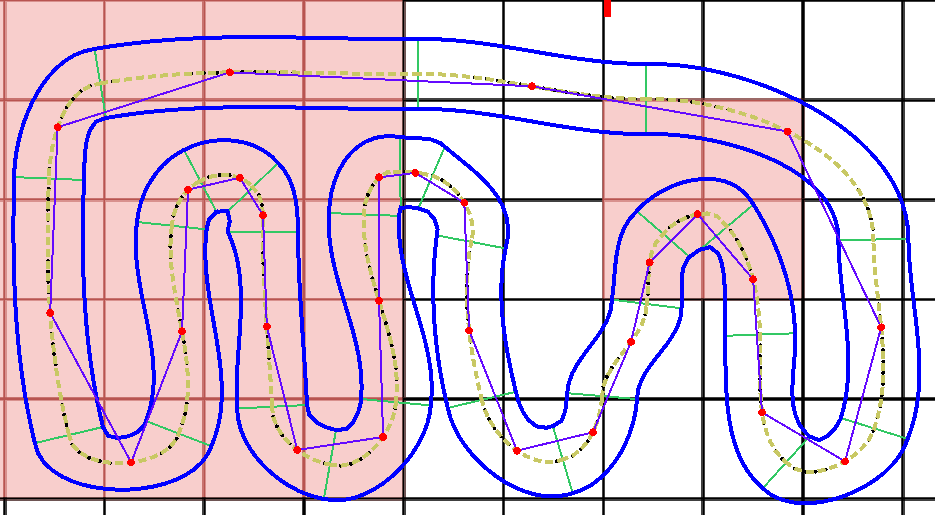}}}\\
\hline\hline\noalign{\smallskip}
Ours w/o map  & $20.47 \pm 0.91$ & $61.32 \pm 14.04$ & $10.09 \pm 14.06$ \\
Ours (AIS)  & $\mathbf{19.08 \pm 0.89}$ & $\mathbf{56.14 \pm 8.17}$ & $\mathbf{8.78 \pm 10.82}$ \\
\end{tabular}
\caption{Quantitative evaluation of the driving performance of our approach. The results show the efficiency of our approach, which significantly reduces the cross-track error while achieving faster lap times in comparison to a model without a mapping network.}
\label{drift:tab:quantitative_driving}
\vspace{-0.5cm}
\end{table}

We observe that latent mapping helps in particular to improve the violation of lane boundaries as our model prevents unexpected drifting that leads off the racing track.
We illustrate such a scenario in Fig.~\ref{drift:fig:path_overlay}.
Further, actions obtained from optimizing erroneous trajectory estimates sometimes led to failure cases on slippery surfaces. In these cases, we had to manually intervene to prevent damage to the car. This happened three times for the model without map updates, while we never observed it for the model with map updates.
This behavior indicates that on slippery surfaces the model with latent surface information predicts the dynamics more accurately while the model without overestimates the speed at which certain corners can be taken as it is trained to predict well averaged over all surfaces including those for which such maneuvers are possible.

\vspace{-0.03cm}
\subsection{Progressive Map Building}
\vspace{-0.03cm}
We further investigate the quality of the latent map with respect to the number of driven laps. 
The mapping model can improve its estimate of the underlying latent vector and variance with every new lap.
This experiment is conducted on Track $3$ while driving ten laps. 
Tab.~\ref{drift:tab:quantitative_laps} shows the lap times and the L2 error that is computed by considering the state transitions of all laps but only using the observations of the first ten laps to build the latent map respectively. The results show that with each additional lap our mapping model is consistently improving the latent estimates to improve the state predictions as well as the driving performance.

\begin{table}
\setlength{\tabcolsep}{3pt}
\footnotesize  % original
% \scriptsize  % smaller
\centering
\begin{tabular}{l|c|c|c|c|c|c}
Lap number &  1 &  2 & 3 & 4 & 5 & 10\\
\noalign{\smallskip}\hline\hline\noalign{\smallskip}
Lap time [s] $\downarrow$  & $20.89$ & $20.62$ & $20.08$ & $19.44$ & $19.32$ & $\mathbf{19.10}$ \\
\noalign{\smallskip}\hline\noalign{\smallskip}
$L2_{10}$ $\downarrow$ & $0.0857$  & $0.0816$ & $0.0807$ & $0.0805$ & $0.0804$  & $\mathbf{0.0801}$\\
\noalign{\smallskip}\hline\noalign{\smallskip}
\end{tabular}
\caption{Quantitative evaluation of the influence of completed laps with respect to the dynamics prediction accuracy and lap time. In this experiment, we only consider map updates during a specified number of consecutive laps. The respective L2 error is calculated over the state transitions of all laps. }
\label{drift:tab:quantitative_laps}
\vspace{-0.5cm}
\end{table}

%\vspace{-0.05cm}
\section{CONCLUSION}
\vspace{-0.05cm}
\label{drift:sec:conclusion}

In this paper, we present a novel approach to estimate latent representations of the road surface. The latent representation is aggregated to a global grid map, which is then leveraged to improve future state predictions from a dynamics model.
To this end, an ensemble of probabilistic dynamics models estimates the distribution of the next state given the current state, current action, and the latent representation of the road that corresponds to the current position of the vehicle. Thus, our dynamics model can effectively leverage the latent cues making it surface-aware. We show the efficacy of our method on a newly proposed dataset, presenting improved prediction accuracy of future states. Furthermore, we employ a model predictive controller and a novel reward function. In real-world driving experiments we demonstrate that including the latent mapping provides a significantly improved driving performance, reducing lap times and enhancing vehicle safety on surfaces with varying friction.
While this work considers the use of surface information to improve the dynamics model, there are also other changing factors influencing the dynamics such as tire pressure and temperature. 
A promising avenue of future work is to integrate our approach in a more complete system with additional sensors to also take these factors into account.

%\addtolength{\textheight}{-12cm}   % This command serves to balance the column lengths
                                  % on the last page of the document manually. It shortens
                                  % the textheight of the last page by a suitable amount.
                                  % This command does not take effect until the next page
                                  % so it should come on the page before the last. Make
                                  % sure that you do not shorten the textheight too much.

%%%%%%%%%%%%%%%%%%%%%%%%%%%%%%%%%%%%%%%%%%%%%%%%%%%%%%%%%%%%%%%%%%%%%%%%%%%%%%%%

% \vspace{-0.5cm}
{\small
\bibliographystyle{IEEEtran}
\bibliography{egbib}
}

\end{document}